\documentclass[11pt,a4paper]{article}
\usepackage[hyperref]{emnlp2020}
\usepackage{times}
\usepackage{latexsym}

\usepackage{microtype}

\aclfinalcopy %
\def\aclpaperid{1574} %

\usepackage{hyperref}
\usepackage{graphicx}
\usepackage{subfigure}
\usepackage{xspace}
\usepackage{booktabs} %
\usepackage{amsmath}
\usepackage{amssymb}
\usepackage{multirow}
\usepackage{subfigure}
\usepackage{bbm}
\usepackage{nicefrac}

\usepackage{pifont}%
\newcommand{\xmark}{\ding{55}\xspace}
\newcommand{\cmark}{\ding{51}\xspace}

\usepackage{color}
\newcommand{\ie}{i.e.,\xspace}
\newcommand{\eg}{e.g.,\xspace}
\newcommand{\eat}[1]{}

\newcommand{\baby}{Theseus Compression\xspace}
\newcommand{\bertot}{BERT-of-Theseus\xspace}

\title{\bertot: Compressing BERT by Progressive Module Replacing}

  \author{Canwen Xu$^1$\thanks{\ \ \ Equal contribution. Work done during these two authors’ internship at Microsoft Research Asia.}, Wangchunshu Zhou$^2$$^*$, Tao Ge$^3$, Furu Wei$^3$, Ming Zhou$^3$\\
 $^1$ University of California, San Diego $^2$ Beihang University $^3$ Microsoft Research Asia \\
 $^1$ {\tt cxu@ucsd.edu} $^2$ {\tt zhouwangchunshu@buaa.edu.cn} \\ $^3$ {\tt \{tage,fuwei,mingzhou\}@microsoft.com}
}

\date{}

\begin{document}
\maketitle
\begin{abstract}
In this paper, we propose a novel model compression approach to effectively compress BERT by progressive module replacing. Our approach first divides the original BERT into several modules and builds their compact substitutes. Then, we randomly replace the original modules with their substitutes to train the compact modules to mimic the behavior of the original modules. We progressively increase the probability of replacement through the training. In this way, our approach brings a deeper level of interaction between the original and compact models. Compared to the previous knowledge distillation approaches for BERT compression, our approach does not introduce any additional loss function. Our approach outperforms existing knowledge distillation approaches on GLUE benchmark, showing a new perspective of model compression.\footnote{The code and pretrained model are available at \url{https://github.com/JetRunner/BERT-of-Theseus}}
\end{abstract}

\section{Introduction}

With the prevalence of deep learning, many huge neural models have been proposed and achieve state-of-the-art performance in various fields~\cite{resnet,transformer}. Specifically, in Natural Language Processing (NLP), pretraining and fine-tuning have become the new norm of most tasks. Transformer-based pretrained models~\cite{bert,roberta,xlnet,mass,unilm} have dominated the field of both Natural Language Understanding (NLU) and Natural Language Generation (NLG).  These models benefit from their ``overparameterized'' nature~\cite{doubledescent} and contain millions or even billions of parameters, making it computationally expensive and inefficient considering both memory consumption and high latency. This drawback enormously hinders the applications of these models in production. 

To resolve this problem, many techniques have been proposed to compress a neural network. Generally, these techniques can be categorized into Quantization~\cite{quantization}, Weights Pruning~\cite{pruning} and Knowledge Distillation (KD)~\cite{kd}. Among them, KD has received much attention for compressing pretrained language models. KD exploits a large \textit{teacher} model to ``teach'' a compact \textit{student} model to mimic the teacher's behavior. In this way, the knowledge embedded in the teacher model can be transferred into the smaller model. However, the retained performance of the student model relies on a well-designed distillation loss function which forces the student model to behave as the teacher. Recent studies on KD~\cite{pkd,tinybert} even leverage more sophisticated model-specific distillation loss functions for better performance.

\begin{figure*}
\centering 
 \subfigure[Compression Training]{ 
    \label{fig:replace:a} %
    \includegraphics[height=2.2in]{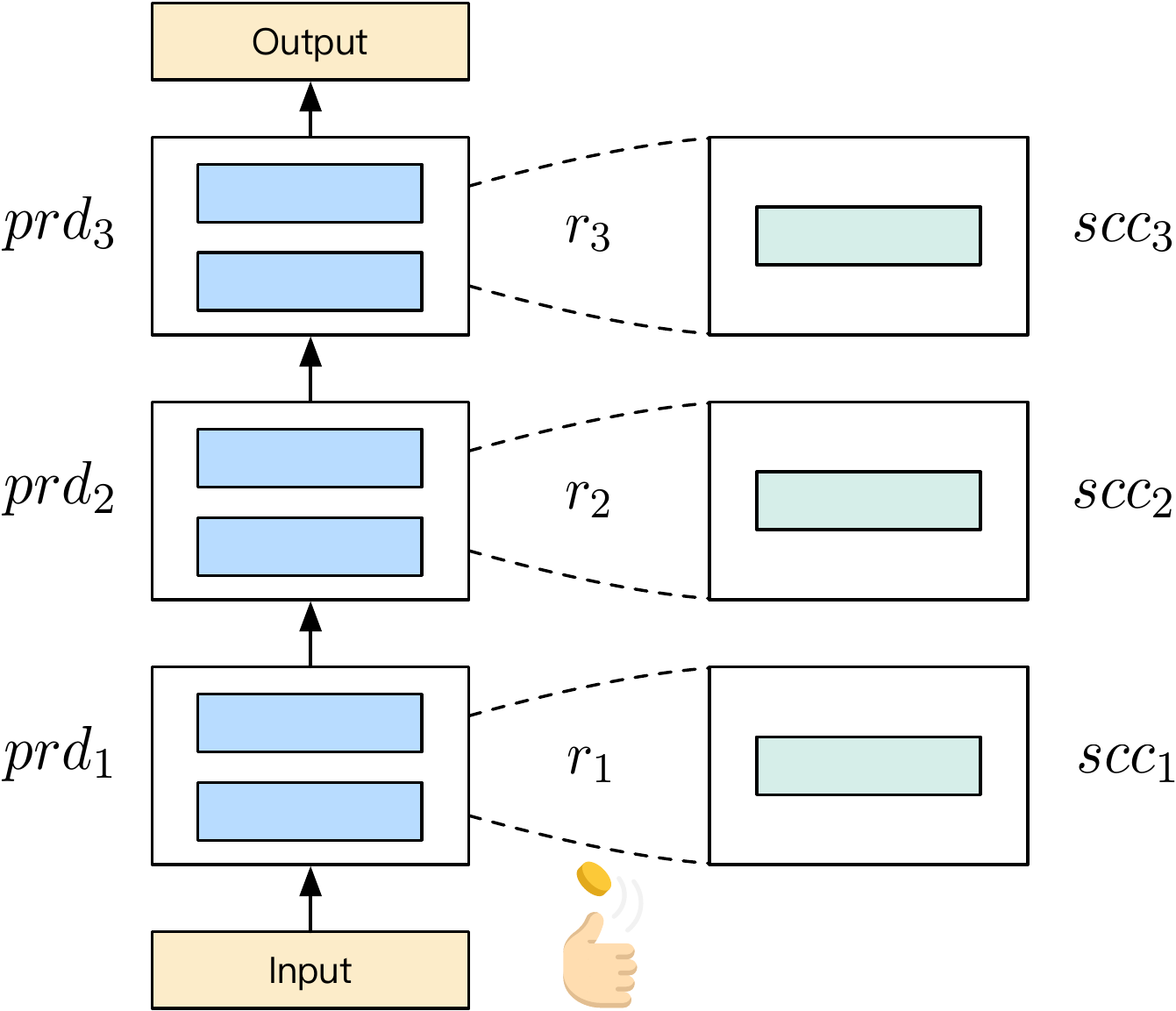}
  } 
  \hspace{1in} 
  \subfigure[Successor Fine-tuning and Inference]{ 
    \label{fig:replace:b} %
    \includegraphics[height=2.2in]{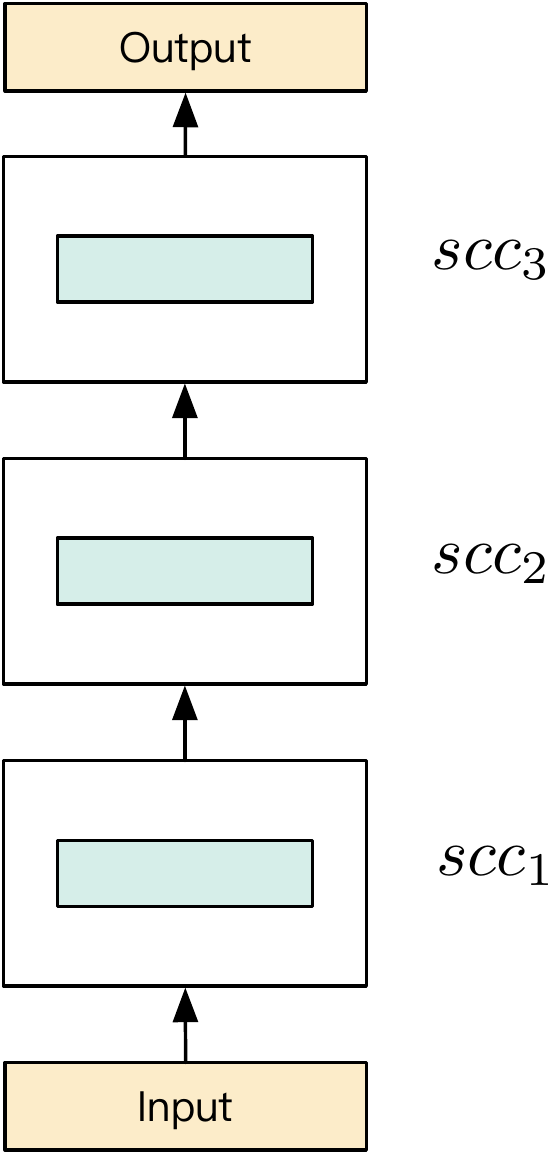}
   } 
  \caption{The workflow of \bertot. In this example, we compress a 6-layer predecessor $P=\{prd_1, \ldots, prd_3\}$ to a 3-layer successor $S=\{scc_1, \ldots, scc_3\}$. $prd_i$ and $scc_i$ contain two and one layer, respectively. (a) During module replacing training, each predecessor module $prd_i$ is replaced with corresponding successor module $scc_i$ by the probability of $p$. (b) During successor fine-tuning and inference, all successor modules $scc_{1\ldots3}$ are combined for calculation.} 
  \label{fig:replace} %
\end{figure*}

Different from previous KD studies which explicitly exploit a distillation loss to minimize the distance between the teacher model and the student model, we propose a new genre of model compression. Inspired by the famous thought experiment ``Ship of Theseus''\footnote{\url{https://en.wikipedia.org/wiki/Ship_of_Theseus}} in Philosophy, where all components of a ship are gradually replaced by new ones until no original component exists, we propose \textbf{Theseus Compression} for BERT (\textbf{\bertot}), which progressively substitutes modules of BERT with modules of fewer parameters. We call the original model and compressed model \textit{predecessor} and \textit{successor}, in correspondence to the concepts of \textit{teacher} and \textit{student} in KD, respectively. As shown in Figure \ref{fig:replace}, we first specify a substitute (successor module) for each predecessor module (\ie modules in the predecessor model). Then, we randomly replace each predecessor module with its corresponding successor module by a probability and make them work together in the training phase. After convergence, we combine all successor modules to be the successor model for inference. In this way, the large predecessor model can be compressed into a compact successor model.

\baby shares a similar idea with KD, which encourages the compressed model to behave like the original, but holds many merits. First, we only use the task-specific loss function in the compression process. However, KD-based methods use task-specific loss, together with one or multiple distillation losses as its optimization objective. Also, selecting various loss functions and balancing the weights of each loss for different tasks and datasets can be laborious~\cite{pkd,distilbert}. Second, different from recent work~\cite{tinybert}, \baby does not use Transformer-specific features for compression thus is potential to compress a wide spectrum of models. Third, instead of using the original model only for inference in KD, our approach allows the predecessor model to work in association with the compressed successor model, enabling a possible gradient-level interaction. Moreover, the different module permutations mixing both predecessor and successor modules may add extra regularization, similar to Dropout~\cite{dropout}. With a Curriculum Learning~\cite{curriculum} driven replacement scheduler, our approach achieves promising performance compressing BERT~\cite{bert}, a large pretrained Transformer model. 

To summarize, our contribution is two-fold: (1) We propose a novel approach, \baby, revealing a new pathway to model compression, with no additional loss function. (2) Our compressed BERT model is $1.94\times$ faster while retaining more than $98\%$ performance of the original model, outperforming other KD-based compression baselines.

\section{Related Work}
\paragraph{Model Compression} Model compression aims to reduce the size and computational cost of a large model while retaining as much performance as possible. Conventional explanations~\cite{nips13:denil,nips16:zhai} claim that the large number of weights is necessary for the training of deep neural network but a high degree of redundancy exists after training. Recent work~\cite{lth} proposes The Lottery Ticket Hypothesis claiming that dense, randomly initialized and feed-forward networks contain subnetworks that can be recognized and trained to get a comparable test accuracy to the original network. Quantization~\cite{quantization} reduces the number of bits used to represent a number in a model. Weights Pruning~\cite{pruning,iccv17:he} conducts a binary classification to decide which weights to be trimmed from the model. Knowledge Distillation (KD)~\cite{kd} aims to train a compact model which behaves like the original one. FitNets~\cite{fitnets} demonstrates that ``hints'' learned by the large model can benefit the distillation process. Born-Again Neural Network~\cite{ban} reveals that ensembling multiple identical-parameterized students can outperform a teacher model.  LIT~\cite{lit} introduces block-wise intermediate representation training. \citet{liu2019improving} distilled knowledge from ensemble models to improve the performance of a single model on NLU tasks. \citet{iclr19:xuta} exploited KD for multi-lingual machine translation. Different from KD-based methods, our proposed \baby is the first approach to mix the original model and compact model for training. Also, no additional loss is used throughout the whole compression procedure, which simplifies the implementation.

\paragraph{Faster BERT} Very recently, many attempts have been made to speed up a large pretrained language model (\eg BERT~\cite{bert}). \citet{betterthanone} reduced the parameters of a BERT model by pruning unnecessary heads in the Transformer. \citet{qbert} quantized BERT to 2-bit using Hessian information. Also, substantial modification has been made to Transformer architecture. \citet{layerdrop} exploited a structure dropping mechanism to train a BERT-like model which is resilient to pruning. ALBERT~\cite{albert} leverages matrix decomposition and parameter sharing. However, these models cannot exploit ready-made model weights and require a full retraining. \citet{distillbilstm} used a BiLSTM architecture to extract task-specific knowledge from BERT. DistilBERT~\cite{distilbert} applies a naive Knowledge Distillation on the same corpus used to pretrain BERT. Patient Knowledge Distillation (PKD)~\cite{pkd} designs multiple distillation losses between the module hidden states of the teacher and student models. Pretrained Distillation~\cite{pd} pretrains the student model with a self-supervised masked LM objective on a large corpus first, then performs a standard KD on supervised tasks. TinyBERT~\cite{tinybert} conducts the Knowledge Distillation twice with data augmentation. MobileBERT~\cite{mobilebert} devises a more computationally efficient architecture and applies knowledge distillation with a bottom-to-top layer training procedure. PABEE~\cite{zhou2020bert} exploits early exiting to dynamically accelerate the inference of BERT.

\section{\bertot}
In this section, we introduce module replacing, the technique proposed for \bertot. Further, we introduce a Curriculum Learning driven scheduler to obtain better performance. The workflow is shown in Figure \ref{fig:replace}.

\subsection{Module Replacing}
The basic idea of \baby is very similar to KD. We want the successor model to act like a predecessor model. KD explicitly defines a loss to measure the similarity of the teacher and student. However, the performance vastly relies on the design of the loss function~\cite{kd,pkd,tinybert}. This loss function needs to be combined with task-specific loss~\cite{pkd,lit}. Different from KD, \baby only requires one task-specific loss function (\eg Cross Entropy), which closely resembles a fine-tuning procedure. Inspired by Dropout~\cite{dropout}, we propose module replacing, a novel technique for model compression. We call the original model and the target model \textit{predecessor} and \textit{successor}, respectively. First, we specify a successor module for each module in the predecessor. For example, in the context of BERT compression, we let one Transformer layer be the successor module for two Transformer layers. Consider a predecessor model $P$ which has $n$ modules and a successor model $S$ which has $n$ predefined modules. Let $P=\{prd_1, \ldots, prd_n\}$ denote the predecessor model, $prd_i$ and $scc_i$ denote the the predecessor modules and their corresponding substitutes, respectively. The output vectors of the $i$-th module is denoted as $\mathbf{y}_i$. Thus, the forward operation can be described in the form of:
\begin{equation}
    \mathbf{y}_{i+1} = prd_i(\mathbf{y}_i)
\end{equation}

During compression, we apply module replacing. First, for $(i+1)$-th module, $r_{i+1}$ is an independent Bernoulli random variable which has probability $p$ to be $1$ and $1-p$ to be $0$.
\begin{equation}
    r_{i+1} \sim \operatorname{Bernoulli}(p)
\end{equation}
Then, the output of the $(i+1)$-th model is calculated as:
\begin{equation}
    y_{i+1} = r_{i+1} * scc_i(\mathbf{y}_i) + (1 - r_{i+1}) * prd_i(\mathbf{y}_i)
\end{equation}
where $*$ denotes the element-wise multiplication, $r_{i+1}\in \{0, 1\}$. In this way, the predecessor modules and successor modules work together in the training. Since the permutation of the hybrid model is random, it adds extra noises as a regularization for the training of the successor, similar to Dropout~\cite{dropout}.

During training, similar to a fine-tuning process, we optimize a regular task-specific loss, \eg Cross Entropy:
\begin{equation}
\label{loss}
	L = -\sum_{j \in \left|X\right|}\sum_{c \in C}\left[\mathbbm{1}\left[\mathbf{z}_{j}=c\right]\cdot\log P\left(\mathbf{z}_j=c|\mathbf{x}_j\right)\right]
\end{equation}
where $\mathbf{x}_j \in X$ is the $i$-th training sample; $\mathbf{z}_j$ is its corresponding ground-truth label; $c$ and $C$ denote a class label and the set of class labels, respectively. For back-propagation, the weights of all predecessor modules are frozen. For both the embedding layer and output layer of the predecessor model are weight-frozen and directly adopted for the successor model in this training phase. In this way, the gradient can be calculated across both the predecessor and successor modules, allowing deeper interaction.

\subsection{Successor Fine-tuning and Inference}
\label{sec:sft}
To make the training and inference processes as close as possible, we further carry out a post-replacement fine-tuning phase to allow all successor modules to work together.
After the replacing compression converges, we collect all successor modules and combine them to be the successor model $S$:
\begin{equation}
\label{equ:inf}
\begin{aligned}
	&S =\{scc_1, \ldots, scc_n\} \\
    &\mathbf{y}_{i+1} = scc_i(\mathbf{y}_i)
\end{aligned}
\end{equation}
Since each $scc_i$ is smaller than $prd_i$ in size, the predecessor model $P$ is in essence compressed into a smaller model $S$.
Then, we fine-tune the successor model by optimizing the same loss of Equation~\ref{loss}. The whole procedure including module replacing and successor fine-tuning is illustrated in Figure \ref{fig:curve}(a). Finally, we use the fine-tuned successor for inference as Equation~\ref{equ:inf}.

\subsection{Curriculum Replacement}
\label{subsec:curriculum}

\begin{figure} 
\centering 
\includegraphics[width=\columnwidth]{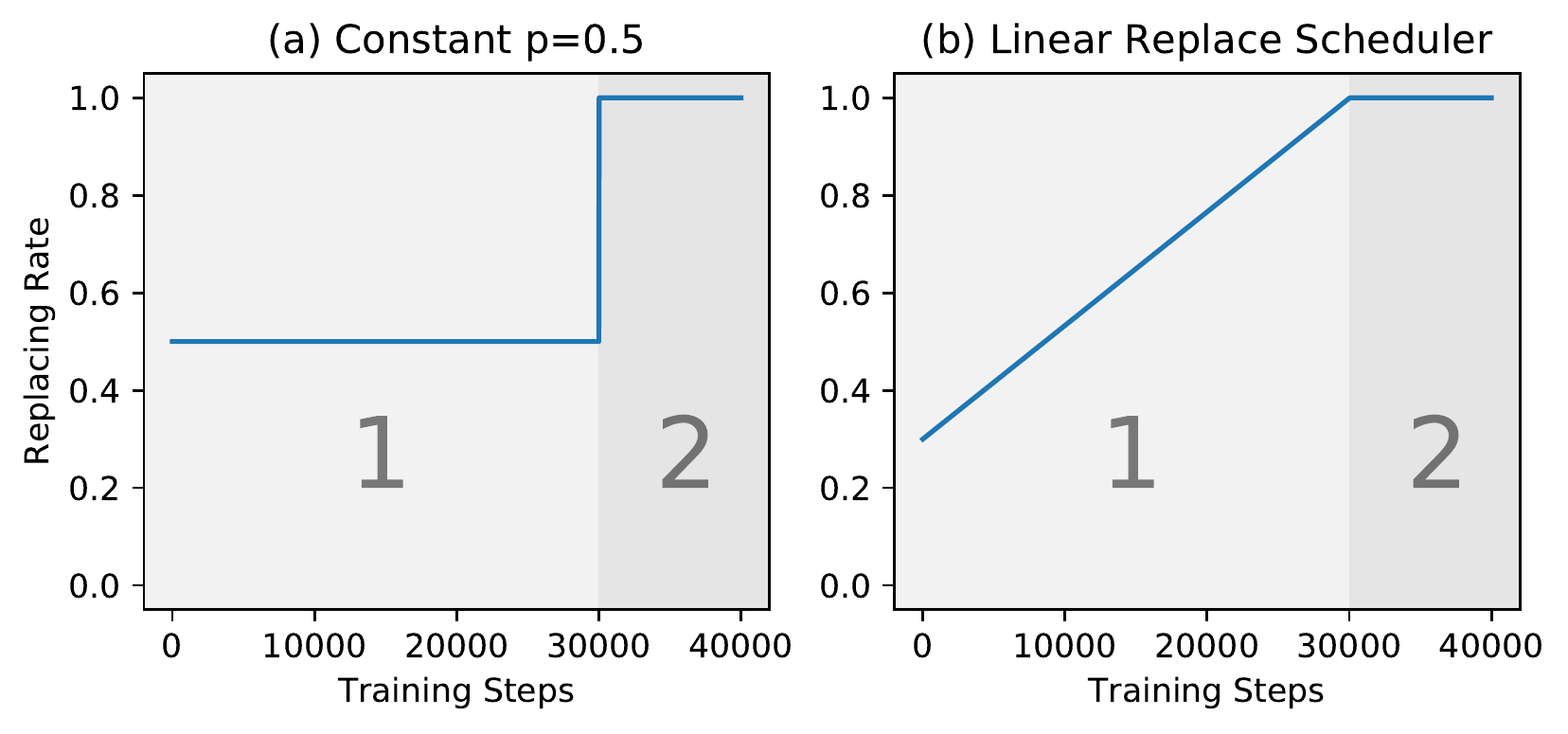}

 \caption{\label{fig:curve}The replacing curves of a constant module replace rate and a replacement scheduler. We use different shades of gray to mark the two phases of \baby: (1) Module replacing. (2) Successor fine-tuning.} 
\end{figure}

Although setting a constant replacement rate $p$ can meet the need for compressing a model, we further highlight a Curriculum Learning~\cite{curriculum} driven replacement scheduler, which coordinates the progressive replacement of the modules. Similar to \cite{curriculumdropout,zhou2020scheduled}, we devise a replacement scheduler to dynamically tune the replacement rate $p$. 

Here, we leverage a simple linear scheduler $\theta(t)$ to output the dynamic replacement rate $p_d$ for step $t$.
\begin{equation}
\label{equ:curriculum}
	p_d = \min(1, \theta(t)) = \min(1, kt + b)
\end{equation}
where $k > 0$ is the coefficient and $b$ is the basic replacement rate. The replacing rate curve with a replacement scheduler is illustrated in Figure~\ref{fig:curve}(b).

In this way, we unify the two previously separated training stages and encourage an end-to-end easy-to-hard learning process. First, with more predecessor modules present, the model would more likely to correctly predict thus have a relatively small cross-entropy loss, which is helpful for smoothing the learning process. Then, at a later time of compression, more modules can be present together, encouraging the model to gradually learn to predict with less guidance from the predecessor and steadily transit to the successor fine-tuning stage.

Second, at the beginning of the compression, when $\theta(t)<1$, considering the average learning rate for all $n$ successor modules, the expected number of replaced modules is $n\cdot p_d$ and the expected average learning rate is:
\begin{equation}
	lr' = (np_d/n)lr = (kt+b)lr
\end{equation}
where $lr$ is the constant learning rate set for the compression and $lr'$ is the equivalent learning rate considering all successor modules. Thus, when applying a replacement scheduler, a warm-up mechanism~\cite{transformertips} is essentially adopted at the same time, which helps the training of a Transformer.

\section{Experiments}
In this section, we introduce the experiments of \baby for BERT~\cite{bert} compression. We compare \bertot with other compression methods and further conduct experiments to analyze the results.

\begin{table*}[t]
\vskip 0.1in
\begin{center}
\small
\resizebox{\textwidth}{!}{
\begin{tabular}{lcclll}
\toprule
Method & \# Layer & \# Param. & Loss Function & External Data Used? & Model-Agnostic? \\
\midrule
BERT-base~\shortcite{bert} & 12 & 110M & CE\textsubscript{MLM} + CE\textsubscript{NSP} & - & - \\
\midrule
Fine-tuning & 6 & 66M & CE\textsubscript{TASK} & \xmark & \cmark \\
Vanilla KD~\shortcite{kd} & 6 & 66M & CE\textsubscript{KD} + CE\textsubscript{TASK} & \xmark & \cmark \\
BERT-PKD~\shortcite{pkd} & 6 & 66M & CE\textsubscript{KD} + PT\textsubscript{KD} + CE\textsubscript{TASK} & \xmark & \cmark \\
\midrule
DistilBERT~\shortcite{distilbert} & 6 & 66M & CE\textsubscript{KD} + Cos\textsubscript{KD} + CE\textsubscript{MLM} & \cmark (unlabeled) & \cmark \\
PD-BERT~\shortcite{pd} & 6 & 66M & CE\textsubscript{MLM} + CE\textsubscript{KD} + CE\textsubscript{TASK} & \cmark (unlabeled) & \cmark \\
\midrule
TinyBERT~\shortcite{tinybert} & 4 & 15M & MSE\textsubscript{attn} + MSE\textsubscript{hidn} + MSE\textsubscript{embd} + CE\textsubscript{KD} & \cmark (unlabeled + labeled) & \xmark\\
MobileBERT~\shortcite{mobilebert} & 24 & 25M & FMT+AT+PKT+CE\textsubscript{KD}+CE\textsubscript{MLM} & \cmark(unlabeled) & \xmark \\
\midrule
\bertot (Ours) & 6 & 66M & CE\textsubscript{TASK} & \xmark & \cmark \\
\bottomrule
\end{tabular}
}
\caption{Comparison of different BERT compression approaches. ``CE'' and ``MSE'' stand for Cross Entropy and Mean Square Error, respectively. ``KD'' indicates the loss is for Knowledge Distillation. ``CE\textsubscript{TASK}'', ``CE\textsubscript{MLM}'' and ``CE\textsubscript{NSP}'' indicate Cross Entropy calculated on downstream tasks, Masked LM pretraining and Next Sentence Prediction, respectively. Other loss functions are described in their corresponding papers.}
\label{tab:baseline}
\end{center}
\vskip -0.1in
\end{table*}

\subsection{Datasets}
We evaluate our proposed approach on the GLUE benchmark~\cite{glue,mrpc,senteval,sst,mnli,qnli,cola}. 
Note that we exclude WNLI~\cite{wnli} following the original BERT paper~\cite{bert}.

The accuracy is used as the metric for SST-2, MNLI-m, MNLI-mm, QNLI and RTE. The F1 and accuracy are used for MRPC and QQP. The Pearson correlation and Spearman correlation are used for STS-B. Matthew's correlation is used for CoLA. The results reported for the test set of GLUE are in the same format as on the official leaderboard. For the sake of comparison with \cite{distilbert}, on the development set of GLUE, the result of MNLI is an average on MNLI-m and MNLI-mm; the results on MRPC and QQP are reported with the average of F1 and accuracy; the result reported on STS-B is the average of the Pearson and Spearman correlation.

\subsection{Experimental Settings}
We test our approach under a task-specific compression setting~\cite{pkd,pd} instead of a pretraining compression setting~\cite{distilbert,mobilebert}. That is to say, we use no external unlabeled corpus but only the training set of each task in GLUE to compress the model. The reason behind this decision is that we intend to straightforwardly verify the effectiveness of our generic compression approach. The fast training process of task-specific compression (\eg no longer than $20$ GPU hours for any task of GLUE) computationally enables us to conduct more analytical experiments. For comparison, DistilBERT~\cite{distilbert} takes $720$ GPU hours to train.  Plus, in real-world applications, this setting provides with more flexibility when selecting from different pretrained LMs (\eg BERT, RoBERTa~\cite{roberta}) for various downstream tasks and it is easy to adopt a newly released model, without a time-consuming pretraining compression. We will also discuss the possibility to use an MNLI model for a general purpose with intermediate transfer learning~\cite{intermediate}.

Formally, we define the task of compression as trying to retain as much performance as possible when compressing the officially released BERT-base (uncased)\footnote{\url{https://github.com/google-research/bert}} to a 6-layer compact model with the same hidden size, following the settings in \cite{distilbert,pkd,pd}. Under this setting, the compressed model has 24M parameters for the token embedding (identical to the original model) and 42M parameters for the Transformer layers and obtains a $1.94\times$ speed-up for inference.

\subsection{Training Details}
We fine-tune BERT-base as the predecessor model for each task with the batch size of $32$, the learning rate of $2 \times 10^{-5}$, and the number of epochs as $4$. As a result, we are able to obtain a predecessor model with comparable performance with that reported in previous studies \cite{distilbert,pkd,tinybert}.

Afterward, for training successor models, following \cite{distilbert,pkd}, we use the first $6$ layers of BERT-base to initialize the successor model since the over-parameterized nature of Transformer~\cite{transformer} could cause the model unable to converge while training on small datasets. During module replacing, We fix the batch size as 32 for all evaluated tasks to reduce the search space. All $r$ variables only sample once for a training batch. The maximum sequence length is set to 256 on QNLI and 128 for the other tasks. We perform grid search over the sets of learning rate $lr$ as \{1e-5, 2e-5\}, the basic replacing rate $b$ as \{0.1, 0.3\}, the scheduler coefficient $k$ making the dynamic replacing rate increase to $1$ within the first \{1000, 5000, 10000, 30000\} training steps. We apply an early stopping mechanism and select the model with the best performance on the development set. We conduct our experiments on a single Nvidia V100 16GB GPU. The peak memory usage is approximately identical to fine-tuning a BERT-base, since there would be at most 12 layers training at the same time. The training time for each task varies depending on the different sizes of training sets. For example, it takes 20 hours to train on MNLI but less than 30 minutes on MRPC.

\begin{table*}[tb]
\vskip 0.1in
\begin{center}
\begin{small}
\begin{tabular}{l|cccccccc|c}
\toprule
\multirow{2}{*}{Method} & CoLA & MNLI & MRPC & QNLI & QQP & RTE & SST-2 & STS-B & Macro  \\
& (8.5K) & (393K) & (3.7K) & (105K) & (364K) & (2.5K) & (67K) & (5.7K) & Score\\
\midrule
BERT-base~\shortcite{bert} & 54.3 & 83.5 & 89.5 & 91.2 & 89.8 & 71.1 & 91.5 & 88.9 & 82.5 \\
\midrule
DistilBERT~\shortcite{distilbert} & 43.6 & 79.0 & 87.5 & 85.3 & 84.9 & 59.9 & 90.7 & 81.2 & 76.5 \\
PD-BERT~\shortcite{pd} & - & \textbf{83.0} & 87.2 & 89.0 & 89.1 & 66.7 & 91.1 & - & -\\
\midrule
Fine-tuning & 43.4 & 80.1 & 86.0 & 86.9 & 87.8 & 62.1 & 89.6 & 81.9 & 77.2 \\
Vanilla KD~\shortcite{kd} & 45.1 & 80.1 & 86.2 & 88.0 & 88.1 & 64.9 & 90.5 & 84.9 & 78.5\\
BERT-PKD~\shortcite{pkd} & 45.5 & 81.3 & 85.7 & 88.4 & 88.4 & 66.5 & 91.3 & 86.2 & 79.2 \\
LayerDrop~\shortcite{layerdrop} & 45.4 & 80.7 & 85.9 & 88.4 & 88.3 & 65.2 & 90.7 & 85.7 & 78.8\\
\bertot & \textbf{51.1} & 82.3 & \textbf{89.0} & \textbf{89.5} & \textbf{89.6} & \textbf{68.2} & \textbf{91.5} & \textbf{88.7} & \textbf{81.2} \\
\bottomrule
\end{tabular}
\caption{Experimental results (median of 5 runs) on the development set of GLUE. The numbers under each dataset indicate the number of training samples. All models listed above (except BERT-base) have 66M parameters, 6 layers and 1.94$\times$ speed-up.}
\label{tab:dev}
\end{small}
\end{center}
\vskip -0.1in
\end{table*}

\begin{table*}[tb]
\vskip 0.1in
\begin{center}
\begin{small}
\resizebox{\textwidth}{!}{
\begin{tabular}{l|cccccccc|c}
\toprule
\multirow{2}{*}{Method} & CoLA & MNLI-m/mm & MRPC & QNLI & QQP & RTE & SST-2 & STS-B & Macro \\
& (8.5K) & (393K) & (3.7K) & (105K) & (364K) & (2.5K) & (67K) & (5.7K) & Score \\
\midrule
BERT-base~\shortcite{bert} & 52.1 & 84.6~/~83.4 & 88.9~/~84.8 & 90.5 & 71.2~/~89.2 & 66.4 & 93.5 & 87.1~/~85.8 & 80.0 \\
\midrule
PD-BERT~\shortcite{pd} & - & \textbf{82.8~/~82.2} & 86.8~/~81.7 & 88.9 & 70.4~/~88.9 & 65.3 & 91.8 & - & - \\
\midrule
Fine-tuning & 41.5 & 80.4~/~79.7 & 85.9~/~80.2 & 86.7 & 69.2~/~88.2 & 63.6 & 90.7 & 82.1~/~80.0 & 75.6 \\
Vanilla KD~\shortcite{kd} & 42.9 & 80.2~/~79.8 & 86.2~/~80.6 & 88.3 & 70.1~/~88.8 & 64.7 & 91.5 & 82.1~/~80.3 & 76.4\\
BERT-PKD~\shortcite{pkd} & 43.5 & 81.5~/~81.0 & 85.0~/~79.9& 89.0 & 70.7~/~88.9 & 65.5 & 92.0 & 83.4~/~81.6 & 77.0 \\
\bertot & \textbf{47.8} & 82.4~/~82.1 & \textbf{87.6~/~83.2} & \textbf{89.6} & \textbf{71.6~/~89.3} & \textbf{66.2} & \textbf{92.2} & \textbf{85.6~/~84.1} & \textbf{78.6}\\
\bottomrule
\end{tabular}
}
\caption{Experimental results on the test set from the GLUE server. All models listed above (except BERT-base) have 66M parameters, 6 layers and 1.94$\times$ speed-up.}
\label{tab:test}
\end{small}
\end{center}
\vskip -0.1in
\end{table*}

\subsection{Baselines}
As shown in Table \ref{tab:baseline}, we compare the layer numbers, parameter numbers, loss function, external data usage and model agnosticism of our proposed approach to existing methods.
We set up a baseline of vanilla Knowledge Distillation~\cite{kd} as in \cite{pkd}. Additionally, we directly fine-tune a truncated 6-layer BERT model (the bottom 6 layers of the original BERT)\footnote{We also tried the top 6 layers and interleaving 6 layers but both perform worse than the bottom 6 layers.} on GLUE tasks to obtain a natural fine-tuning baseline. Under the setting of compressing 12-layer BERT-base to a 6-layer compact model, we choose BERT-PKD~\cite{pkd}, PD-BERT~\cite{pd}, and DistilBERT~\cite{distilbert} as strong baselines. Note that DistilBERT~\cite{distilbert} is not directly comparable here since it uses a pretraining compression setting. Both PD-BERT and DistilBERT use external unlabeled corpus. Additionally, we use LayerDrop~\cite{layerdrop} on BERT weights to prune the model on downstream tasks. We do not include TinyBERT~\cite{tinybert} since it conducts distillation twice and leverages extra augmented data for GLUE tasks. We also exclude MobileBERT~\cite{mobilebert}, due to its redesigned Transformer block and different model size. Besides, in these two studies, the loss functions are not architecture-agnostic thus limit their applications on other types of models.

\subsection{Experimental Results}
We report the experimental results on the development set of GLUE in Table \ref{tab:dev} and submit our predictions to the GLUE test server and obtain the results from the official leaderboard as shown in Table \ref{tab:test}. Note that DistilBERT does not report on the test set. The BERT-base performance reported on GLUE development set is the predecessor fine-tuned by us. The results of BERT-PKD on the development set are reproduced by us using the official implementation. In the original paper of BERT-PKD, the results of CoLA and STS-B on the test set are not reported, thus we reproduce these two results. Fine-tuning and Vanilla KD baselines are both implemented by us. All other results are from the original papers.\footnote{Please note that the reported results of DistilBERT are different across various versions on arXiv. The results here are from \href{https://arxiv.org/pdf/1910.01108v3.pdf}{v3}, which was the newest version when we composed this paper.} The macro scores here are calculated in the same way as the official leaderboard but are not directly comparable with GLUE leaderboard since we exclude WNLI from the calculation.

Overall, our \bertot retains $98.4\%$ and $98.3\%$ of the BERT-base performance on GLUE development set and test set, respectively. On every task of GLUE, our model dramatically outperforms the fine-tuning baseline, indicating that with the same loss function, our proposed approach can effectively transfer knowledge from the predecessor to the successor. Also, our model obviously outperforms the vanilla KD~\cite{kd} and Patient Knowledge Distillation (PKD)~\cite{pkd}, showing its supremacy over the KD-based compression approaches. On MNLI, our model performs better than BERT-PKD but slightly lower than PD-BERT~\cite{pd}. However, PD-BERT exploits an additional corpus which provides much more samples for knowledge transferring. Also, we would like to highlight that on RTE, our model achieves nearly identical performance to BERT-base and on QQP our model even outperforms BERT-base. To analyze, a moderate model size may help generalize and prevent overfitting on downstream tasks. Notably, on both large datasets with more than 350K samples (\eg MNLI and QQP) and small datasets with fewer than 4K samples (\eg MRPC and RTE), our model can consistently achieve good performance, verifying the robustness of our approach.

\begin{table*}[t]
\vskip 0.1in
\begin{center}
\begin{small}
\begin{tabular}{l|ccccccc}
\toprule
Method & MNLI & MRPC & QNLI & QQP & RTE & SST-2 & STS-B \\
\midrule
BERT-base~\shortcite{bert} & 83.5 & 89.5 & 91.2 & 89.8 & 71.1 & 91.5 & 88.9 \\
\midrule
DistilBERT~\shortcite{distilbert} & 79.0 & \textbf{87.5} & 85.3 & 84.9 & 59.9 & 90.7 & 81.2 \\
PD-BERT~\shortcite{pd} & \textbf{83.0} & 87.2 & \textbf{89.0} & \textbf{89.1} & 66.7 & 91.1 & - \\
\midrule
\bertot\textsubscript{MNLI} & 82.1 & \textbf{87.5} & 88.8 & 88.8 & \textbf{70.1} & \textbf{91.8} & \textbf{87.8}\\
\bottomrule
\end{tabular}
\caption{Experimental results of intermediate-task transfer learning on GLUE-dev.}
\label{tab:gp}
\end{small}
\end{center}
\vskip -0.1in
\end{table*}

\subsection{Intermediate-Task Transfer Learning}
Although our approach achieves good performance under a task-specific setting, it requires more computational resources to fine-tune a full-size predecessor than a compact BERT (\eg DistilBERT~\cite{distilbert}).
\citet{intermediate} found that models trained on some datasets can be used for a second-round fine-tuning. Thus, we use MNLI as the intermediate task and release our compressed model by conducting compression on MNLI to facilitate downstream applications. After compression, we fine-tune the successor model on other sentence classification tasks and compare the results with DistilBERT~\cite{distilbert} in Table \ref{tab:gp}. Our model achieves an identical performance on MRPC and outperforms DistilBERT on the other sentence-level tasks. Also, our intermediate-task transfer results also outperform PD-BERT~\cite{pd} on three tasks, indicating that our task-specific model is also competitive for a general purpose through the intermediate-task transfer learning approach.

\newcommand{\replace}[1]{$prd_#1\rightarrow scc_#1$}
\begin{table}[t]
\vskip 0.1in
\begin{center}
\begin{small}
\resizebox{\columnwidth}{!}{
\begin{tabular}{l|lll}
\toprule
Replacement & QNLI$(\Delta)$ & MNLI$(\Delta)$ & QQP$(\Delta)$ \\
\midrule
Predecessor & 91.87 & 84.54 & 89.48 \\
\midrule
\replace{1} & 88.50 (-3.37) & 81.89 (-2.65) & 88.58 (-0.90) \\
\replace{2} & 90.54 (-1.33) & 83.33 (-1.21) & 88.43 (-1.05) \\
\replace{3} & 90.76 (-1.11) & 83.27 (-1.27) & 88.86 (-0.62) \\
\replace{4} & 90.46 (-1.41) & 83.34 (-1.20) & 88.86 (-0.62) \\
\replace{5} & 90.74 (-1.13) & 84.16 (-0.38) & 89.09 (-0.39) \\
\replace{6} & 90.57 (-1.30) & 84.09 (-0.45) & 89.06 (-0.42) \\

\bottomrule
\end{tabular}
}
\caption{Impact of the replacement for different modules on GLUE-dev. \replace{i} indicates the replacement of the $i$-th module from the predecessor.}
\label{tab:replace}
\end{small}
\end{center}
\vskip -0.1in
\end{table}

\begin{table*}[tb]
\vskip 0.1in
\begin{center}
\begin{small}
\resizebox{\textwidth}{!}{
\begin{tabular}{l|llllllll}
\toprule
Strategy & CoLA$(\Delta)$ & MNLI$(\Delta)$ & MRPC$(\Delta)$ & QNLI$(\Delta)$ & QQP$(\Delta)$ & RTE$(\Delta)$ & SST-2$(\Delta)$ & STS-B$(\Delta)$ \\
\midrule
Constant Rate & 44.4 & 81.9 & 87.1 & 88.5 & 88.6 & 66.4 & 90.6 & 88.4 \\
\midrule
Anti-curriculum & 42.8 (-1.6) & 79.8 (-2.1) & 85.6 (-1.5) & 87.8 (-0.7) & 87.6 (-1.0) & 62.4 (-4.0) & 88.8 (-1.8) & 85.4 (-3.0) \\
Curriculum & \textbf{51.1 (+6.7)} & \textbf{82.3 (+0.4)} & \textbf{89.0 (+1.9)} & \textbf{89.5 (+1.0)} & \textbf{89.6 (+1.0)} & \textbf{68.2 (+1.8)} & \textbf{91.5 (+0.9)} & \textbf{88.7 (+0.3)} \\
\bottomrule
\end{tabular}
}
\caption{Comparison of models compressed with a constant replacing rate, a curriculum replacement scheduler and its corresponding anti-curriculum scheduler on GLUE-dev.}
\label{tab:curriculum}
\end{small}
\end{center}
\vskip -0.1in
\end{table*}

\section{Analysis}
In this section, we conduct extensive experiments to analyze our \bertot.

\subsection{Impact of Module Replacement}
As pointed out in previous work \cite{layerdrop}, different layers of a Transformer play imbalanced roles for inference. To explore the effect of different module replacements, we iteratively use one compressed successor module (constant replacing rate, without successor fine-tuning) to replace its corresponding predecessor module on QNLI, MNLI and QQP, as shown in Table \ref{tab:replace}. Our results show that the replacement of the last two modules have limited influence on the overall performance while the replacement of the first module significantly harms the performance. To analyze, the linguistic features are mainly extracted by the first few layers. Therefore, the reduced representation capability becomes the bottleneck for the following layers.

\subsection{Impact of Replacing Rate}

We attempt to adopt different replacing rates on GLUE tasks. First, we fix the batch size to be $32$ and learning rate $lr$ to be $2\times 10^{-5}$ and conduct compression on each task. On the other hand, as we analyzed in Section \ref{subsec:curriculum}, the equivalent learning rate $lr'$ is affected by the replacing rate. To further eliminate the influence of the learning rate, we fix the equivalent learning rate $lr'$ to be $2\times 10^{-5}$ and adjust the learning rate $lr$ for different replacing rates by $lr= lr' / p$.

\begin{figure}[t]
\centering 
 \subfigure[MRPC]{ 
    \label{fig:rr:mrpc} %
    \includegraphics[width=1.42in]{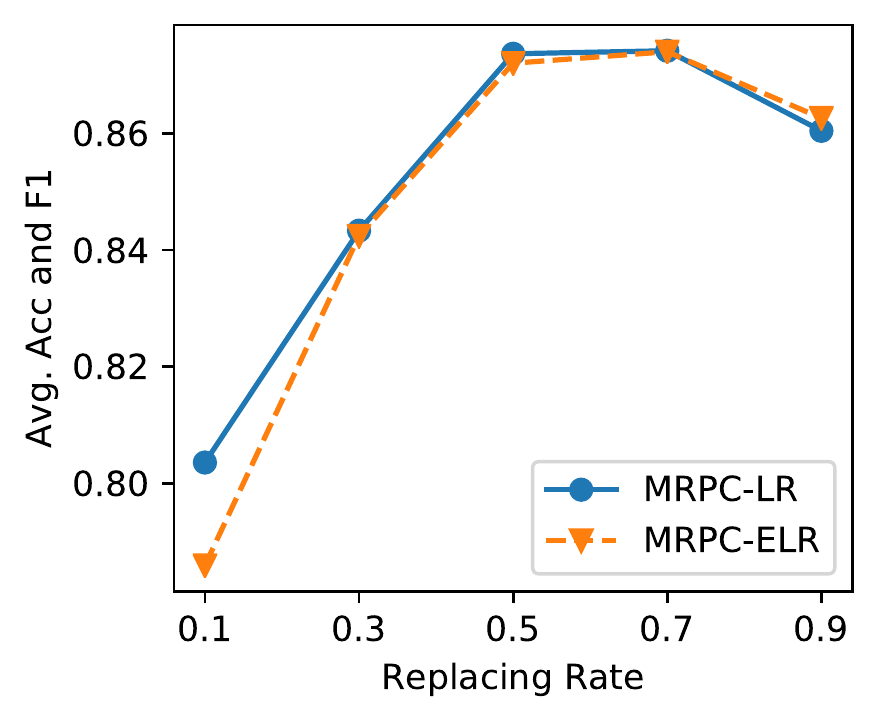}
  } 
  \subfigure[RTE]{ 
    \label{fig:rr:rte} %
    \includegraphics[width=1.42in]{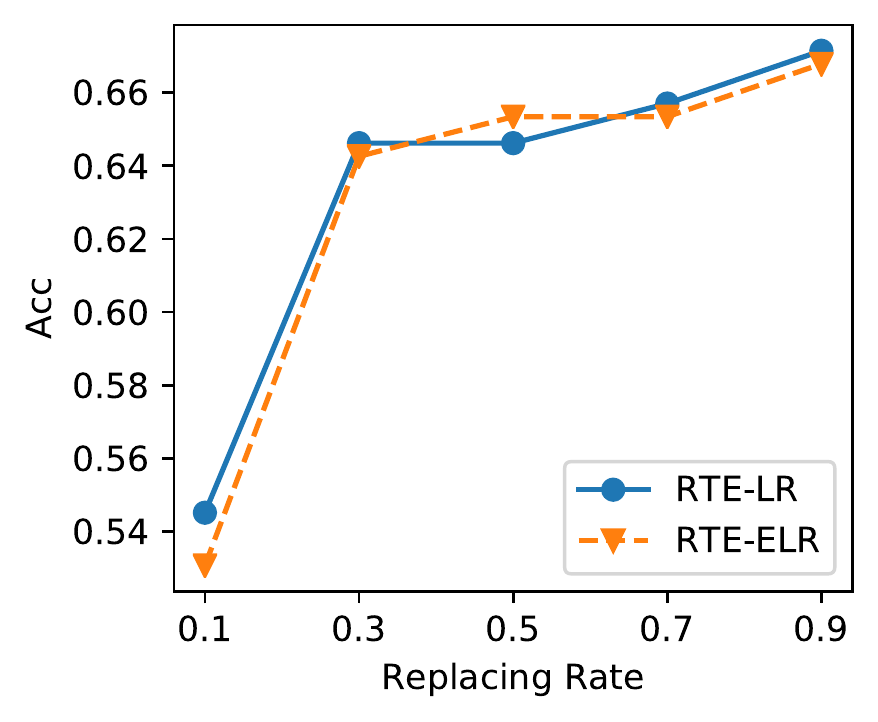}
   } 
  \caption{\label{fig:rr}Performance of different replacing rate on MRPC and RTE. ``LR'' and ``ELR'' denote that the learning rate and equivalent learning rate are fixed, respectively.} %
\end{figure}

We illustrate the results with different replacing rates on two representative tasks (MRPC and RTE) in Figure \ref{fig:rr}. The trivial gap between two curves in both figures indicate that the effect of different replacing rates on equivalent learning rate is not the main factor for the performance differences. A replacing rate in the range between $0.5$ and $0.7$ can always lead to a satisfying performance on all GLUE tasks. However, a significant performance drop can be observed on all tasks if the replacing rate is too small (\eg $p=0.1$). On the other hand, the best replacing rate differs across tasks.

\subsection{Impact of Replacement Scheduler}
To study the impact of our curriculum replacement strategy, we compare the results of \bertot compressed with a constant replacing rate and with a replacement scheduler. The constant replacing rate for the baseline is searched over \{0.5, 0.7, 0.9\}. Additionally, we implement an ``anti-curriculum'' baseline, similar to the one in \cite{curriculumdropout}. For each task, we adopt the same coefficient $k$ and basic replacing rate $b$ to calculate the $p_d$ as Equation \ref{equ:curriculum} for both curriculum replacement and anti-curriculum. However, we use $1-p_d$ as the dynamic replacing rate for anti-curriculum baseline. Thus, we can determine whether the improvement of curriculum replacement is simply due to an inconstant replacing rate or an easy-to-hard curriculum design.

As shown in Table \ref{tab:curriculum}, our model compressed with curriculum scheduler consistently outperforms a model compressed with a constant replacing rate. In contrast, a substantial performance drop is observed on the model compressed with an anti-curriculum scheduler, which further verifies the effectiveness and importance of the curriculum replacement strategy.

\begin{table*}[tb]
\vskip 0.1in
\begin{center}
\begin{small}
\resizebox{\textwidth}{!}{
\begin{tabular}{l|cc|cccccccc|c}
\toprule
\multirow{2}{*}{Method} & \#Layer & Speed- & CoLA & MNLI & MRPC & QNLI & QQP & RTE & SST-2 & STS-B & Macro  \\
& & up & (8.5K) & (393K) & (3.7K) & (105K) & (364K) & (2.5K) & (67K) & (5.7K) & Score\\
\midrule
BERT-base~\shortcite{bert} & 12 & 1.00$\times$ & 54.3 & 83.5 & 89.5 & 91.2 & 89.8 & 71.1 & 91.5 & 88.9 & 82.5 \\
\midrule
Fine-tuning & 6 & 1.94$\times$ & 43.4 & 80.1 & 86.0 & 86.9 & 87.8 & 62.1 & 89.6 & 81.9 & 77.2 \\
\bertot & 6 & 1.94$\times$ & \textbf{51.1} & \textbf{82.3} & \textbf{89.0} & \textbf{89.5} & \textbf{89.6} & \textbf{68.2} & \textbf{91.5} & \textbf{88.7} & \textbf{81.2} \\
\midrule
Fine-tuning & 4 & 2.82$\times$ & 33.9 & 78.4 & 86.0 & 82.3 & 87.1 & 58.2 & 87.2 & 78.4 & 73.9 \\
\bertot & 4 & 2.82$\times$ & \textbf{41.3} & \textbf{80.0} & \textbf{87.5} & \textbf{86.1} & \textbf{88.7} & \textbf{61.9} & \textbf{89.1} & \textbf{82.5} & \textbf{77.2} \\
\midrule
Fine-tuning & 3 & 3.66$\times$ & 27.5 & 78.1 & 81.9 & 80.4 & 86.5 & 57.7 & 85.9 & 76.8 & 71.9 \\
\bertot & 3 & 3.66$\times$ & \textbf{35.0} & \textbf{78.8} & \textbf{84.3} & \textbf{82.1} & \textbf{87.3} & \textbf{59.5} & \textbf{87.2} & \textbf{78.9} & \textbf{74.1} \\
\bottomrule
\end{tabular}
}
\caption{Experimental results of replacing different numbers of layers with one layer on GLUE-dev. ``\#Layer'' indicates the number of layers in the compressed models. }
\label{tab:34}
\end{small}
\end{center}
\vskip -0.1in
\end{table*}

\subsection{Impact of Predecessor Layers}
We further replace different numbers of Transformer layers with one layer to verify the effectiveness of \baby under different settings. We replace 3/4 layers with one Transformer layer, resulting in a 4/3-layer BERT model. The results are shown in Table \ref{tab:34}. \bertot consistently outperforms the fine-tuned truncated BERT baselines, demonstrating its effectiveness under different settings.

\section{Discussion}
In this paper, we propose \baby, a novel model compression approach. We use this approach to compress BERT to a compact model that outperforms other models compressed by Knowledge Distillation. Our work highlights a new genre of model compression and reveals a new path towards model compression.

For future work, we would like to explore the possibility of applying \baby on heterogeneous network modules. First, many developed in-place substitutes (\eg ShuffleNet unit~\cite{shufflenet} for ResBlock~\cite{resnet}, Reformer Layer~\cite{reformer} for Transformer Layer~\cite{transformer}) are natural successor modules that can be directly adopted in \baby. Also, it is possible to use a feed-forward neural network to map features between the hidden spaces of different sizes~\cite{tinybert} to enable replacement between modules with different input and output sizes. Although our model has achieved good performance compressing BERT, it would be interesting to explore its possible applications in other neural models. As summarized in Table \ref{tab:baseline}, our model does not rely on any model-specific features to compress BERT. Therefore, it is potential to apply \baby to other large models (\eg ResNet~\cite{resnet} in Computer Vision). 
In addition, we would like to conduct \baby on more types of neural networks including Convolutional Neural Networks and Graph Neural Networks. We will also investigate the combination of our compression-based approach with recently proposed dynamic acceleration method~\cite{zhou2020bert} to further improve the efficiency of pretrained language models.

\section*{Acknowledgments}
We are grateful for the insightful comments from the anonymous reviewers. Tao Ge is the corresponding author.

\bibliographystyle{acl_natbib}
\bibliography{anthology,emnlp2020}

\end{document}